# POLYNOMIAL TRAJECTORY ALGORITHM FOR A BIPED ROBOT

ERIK CUEVAS[1,2], DANIEL ZALDIVAR[1,2], MARCO PEREZ-CISNEROS[2]

MARTE RAMIREZ[1]

[1]*Institut für Informatik, Freie Universität Berlin*

*Takustr. 9, 14195 Berlin, Germany*

[2]*CUCEI, Universidad de Guadalajara*

*Av. Revolucion 1500,C.P. 44430 Guadalajara, Jal., Mexico*

*{daniel.zaldivar, erik.cuevas, marco.perez}@cucei.udg.mx*

**Abstract**

Building trajectories for biped robot walking is a complex task considering all degrees of freedom (DOFs) commonly bound within the mechanical structure. A typical problem for such robots is the instability produced by violent transitions between walking phases in particular when a swinging leg impacts the surface. Although extensive research on novel efficient walking algorithms has been conducted, falls commonly appear as the walking speed increases or as the terrain condition changes. This paper presents a polynomial trajectory generation algorithm (PTA) to implement the walking on biped robots following the cubic Hermitian polynomial interpolation between initial and final conditions. The proposed algorithm allows smooth transitions between walking phases, significantly reducing the possibility of falling. The algorithm has been successfully tested by generating walking trajectories under different terrain conditions on a biped robot of 10 DOFs. PTA has shown to be simple and suitable to generate real time walking trajectories, despite reduced computing resources of a commercial embedded microcontroller. Experimental evidence and comparisons to other state-of-the-art methods demonstrates a better performance of the proposed method in generating walking trajectories under different ground conditions.

## 1. Introduction

Robots must successfully deal with complicated environments such as rugged terrain, sloped surfaces, and steep stairs. Although it is assumed that biped robots can walk in almost any type of terrain surpassing some of the wheeled robots capabilities [1] [2] [3], they are complex nonlinear systems with many degrees of freedom that may fall down easily while walking due to its relatively small feet and other important design constraints.

A popular technique to implement biped walking is to keep the zero moment point (ZMP) constraint within a supporting feet polygon in order to ensure stable walking gaits [4]. Several methods have been proposed to generate walking trajectories satisfying this condition such as those referred in [5]-[17]. They fall into two groups: time-dependent and time-invariant algorithms. By far, the most popular algorithms are time-dependent which involve the tracking of pre-calculated trajectories. The second group requires the precise knowledge of biped dynamics in order to solve complex nonlinear models to generate walking patterns. This paper focuses only on the time-invariant algorithms.

Several techniques for generating walking motion for biped robots may be found in the literature. For instance, Ohishi et. al, [11] approximated the biped movement as a three dimensional Inverted Pendulum Model (IPM), using the resulting points as tracking reference in Cartesian space. Some other algorithms with small variations of IPM can also be found. In [17] the IPM is transformed into a cart-table model, with the cart movements corresponding to the trajectory of the Center of Mass (CoM). In [12] and [13], Kajita et al, demonstrated the use of a length-varying inverted pendulum as reference point to generate trajectories. The pendulum's length varies as to keep the biped's COM at a constant height above the walking surface (see also [8]). Grishin et al. [14] used a pre-computed trajectory with online adjustment to improve stability of the biped robot. Other works such as those in [5],[6],[7] and [10], use a different inverted pendulum method to generate the walking trajectory. Most of the works on three-dimensional walking movements compute decoupled trajectories from the frontal and sagittal planes.

Katoh and Mori demonstrated in [15] that using a Van der Pol oscillator as generator of the tracking reference would induce walking trajectories for a biped robot. Moreover, Furusho and Masubuchi in [16] presented the walking control algorithms by tracking a piecewise-linear joint reference trajectory.

Another method for trajectory generation is to mimic the human rhythmic function by means of a central pattern generator, just as it is reported in [9]. In this method, one self-oscillating system is designed as to generate synchronized periodic motions for each joint.

Although extensive research on novel efficient walking algorithms has been conducted, the reported results still show a trend for falling down as walking speed increases or when terrain conditions change. Slow down the walking pace will be a temporary solution as the problem remains unsolved as walking deterioration is inflicted on subsequent walking phases. Some previous works [18] have shown the negative effect of a violent impact between the feet and the ground which yields a reaction force $F_R$ and increases the possibility of falls. The effect can be even worse, when the robots' velocity is increased or the terrain conditions change. Although some IPM algorithms have attempted to solve the problem, they require extensive computation unsuitable for real-time applications.

This paper presents the polynomial trajectory algorithm (PTA) which is a simple and effective walking trajectory algorithm based on cubic Hermitian polynomial interpolation of the kinematics positions of the robot. A Hermite spline (HS) is a cubic polynomial interpolation in segments with adjustable derivatives at each control point that allows decreasing link's velocities when they do reach the target point. Therefore, the impact on the leg caused by the ground contact or by violent transitions among different walking phases may also be reduced. The walking trajectory is generated for each joint, adjusting some intermediate positions in order to assure the best ZMP trajectory. The resulting approach generates suitable bipedal walking gaits in real time considering only modest computing resources, such as a microcontroller embedded platform. Experimental evidence shows the effectiveness of the method to generate walking trajectories under different ground conditions.

The proposed algorithm was successfully tested on a biped robot of 10 DOF's [19] under different walking conditions. This paper presents a comparison between several state-of-the-art methods and it demonstrates a better relationship between the link velocities and the reaction forces produced by the proposed method. It reduces the ZMP instability created by violent transitions between walking phases such as the swinging leg impacting the surface.

This paper is organized as follows: Section 2 introduces the biped robot model and the walking gait. Section 3 discusses on the impact model and its velocity discontinuity. Section 4 describes PTA and its parameters while Section 5 features the walking motions generated by the PTA, despite terrain changes. This section also discusses on comparing the PTA performance to other related methods. Finally, Section 6 draws some conclusions.

## 2. Robot model and bipedal walking

### 2.1 Robot description

The dynamics of a biped robot is closely related to its structure [20]. This work employs the "Dany Walker" robot [19], built on 10 low-density aluminum links. Each link consists of a structure which has been carefully designed to allow an effective torque transmission and low deformation. All links are connected within the biped robot structure of 10 degrees of freedom as shown in Fig. 1. The motors allow movements within the frontal and sagittal plane. Figure 2a shows the frontal plane of the robot while the Figure 2b shows the sagittal plane. The embedded control computer system is based on a PIC18F4550 microcontroller executing concurrent tasks such as trajectory generation, servo-motor control, and sensor signal collection. The ZMP is estimated through data integration from pressure sensors (Flexiforce type) located on each foot . The centre of pressure (CoP) matches the ZMP if and only if the latter falls inside the supporting foot polygon (SfP). In the "Dany Walker" biped robot, the ZMP was found by using feedback from triangular force sensor arrangement as it is described in [19,21]. Figure 1 shows the positive direction of the x-axis corresponding to the robot's forward movement. The positive direction of the y-axis corresponds to the robot's movement to its left, while the positive direction of the z-axis is the opposite direction of gravity. Each mass is assumed to be located on the midpoint of its corresponding link. The parameters of the robot model follow the convention presented in [22]. The parameters are summarized in Table 1 following locations drawn in Figure 3.

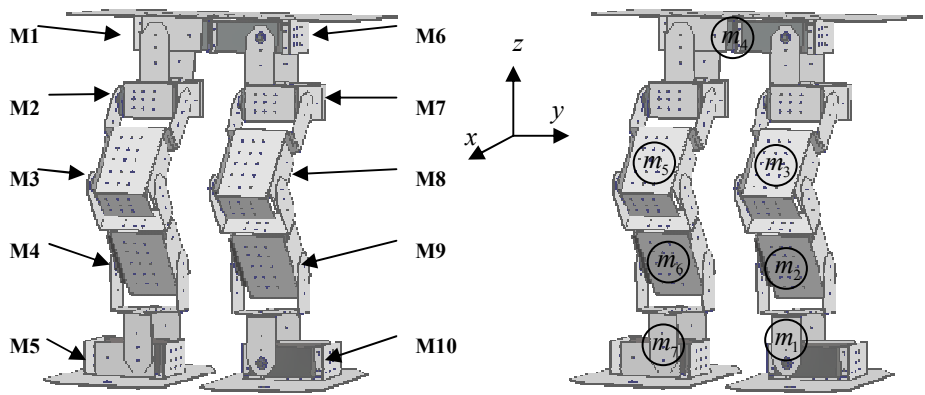

**Fig. 1.** "Dany Walker" biped robot.

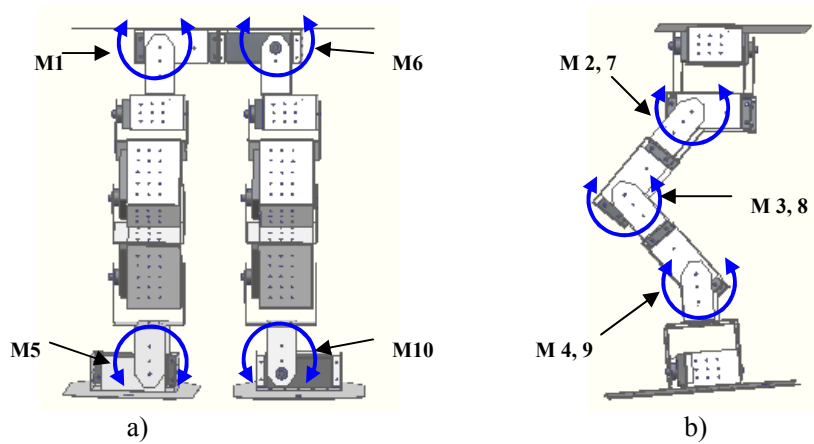

**Fig. 2.** Dany Walker robot a) frontal plane b) sagittal plane.

| Parameter | Value |
|---|---|
| $m_1$ | 0.2Kg |
| $m_2$ | 0.3Kg |
| $m_3$ | 0.3Kg |
| $m_4$ | 0.455Kg |
| $m_5$ | 0.3Kg |
| $m_6$ | 0.3Kg |
| $m_7$ | 0.2Kg |
| $l_1$ | 10cm |
| $l_2$ | 11m |
| $l_3$ | 11cm |
| $l_4$ | 10cm |
| $l_5$ | 11cm |
| $l_6$ | 11cm |
| $l_7$ | 10cm |
| $l_w$ | 11.2cm |
| $l_{fl}$ | 10cm |
| $l_{fw}$ | 8cm |
| $l_{f1}$ | 5cm |
| $l_{f2}$ | 4cm |

**Table 1.** Parameters of the biped robot.

The foot's length and width are 10cm and 8cm respectively. Let position $(x_s, y_s, z_s)$ be middle point of the supporting foot. Thus the robot motion may be expressed with respect to the reference frame whose centre is $(x_s, y_s, z_s)$. The x-directional ZMP ($x_{ZMP}$) and the y-directional ZMP ($y_{ZMP}$) should be located in the following region:

$$-5\text{cm} < x_{ZMP} < 5\text{cm}$$
$$-4\text{cm} < y_{ZMP} < 4\text{cm}.$$
(1)

For the single support phase, such region is a convex hull for all contact points between the foot and the ground. Thus, if the ZMP falls within this area, the biped robot can walk without falling down [23]. However, it is difficult to calculate the ZMP from an 3D robot model as in Figure 1 because of the coupling motion between the frontal plane (y–z plane) and sagittal plane (x–z plane). Therefore, the walking motion may be generated from two 2D models including the sagittal plane model with 8 segments and 7 DoFs as shown in Figure 3(a). The frontal plane model has 6 segments and 5 DoFs as shown by Figure 3(b). The ZMP equations of the sagittal plane robot model (Fig. 3(a)) and the frontal plane robot model (Fig. 3(b)) are:

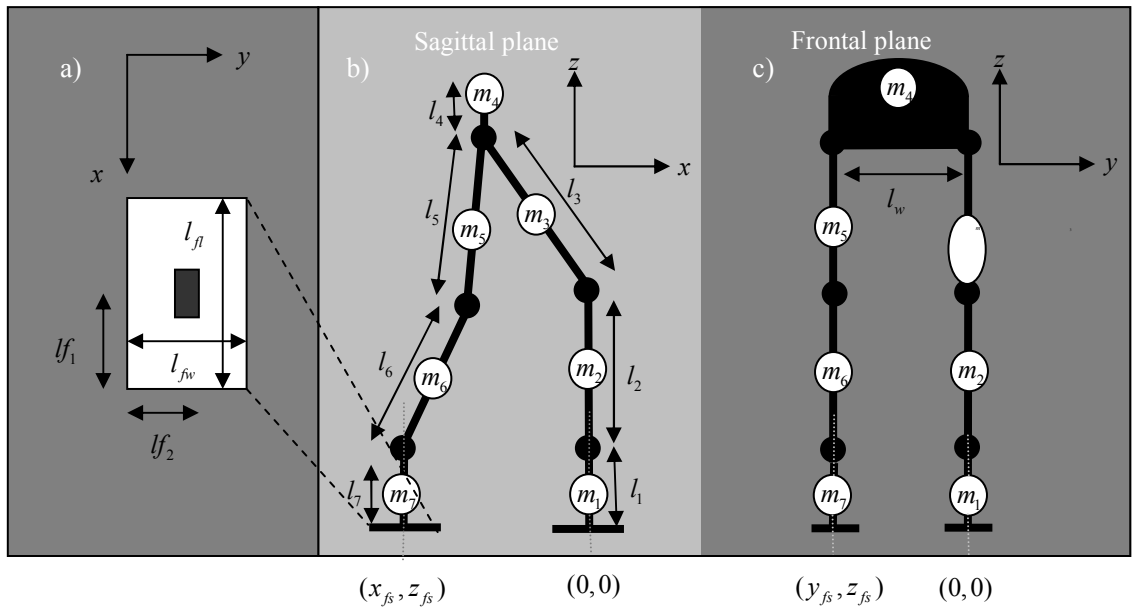

**Fig. 3.** (a) Foot parameters, (b) sagittal plane and (c) frontal plane configuration.

$$x_{ZMP} = \frac{\sum_{i=1}^{7} m_i(\ddot{z}_i + g)x_i - \sum_{i=1}^{7} m_i \ddot{x}_i z_i}{\sum_{i=1}^{7} m_i(\ddot{z}_i + g)} \quad (2)$$

$$y_{ZMP} = \frac{\sum_{i=1}^{7} m_i(\ddot{z}_i + g)y_i - \sum_{i=1}^{7} m_i \ddot{y}_i z_i}{\sum_{i=1}^{7} m_i(\ddot{z}_i + g)} \quad (3)$$

with *g* being the gravity, $(x_i, y_i, z_i)$ and $m_i$ being the position and mass of the *i*-th point mass ($i = 1, \ldots, 7$) [7]. In Equations 2 and 3, the inertia factors can be ignored assuming that the mass of link *i* is uniformly distributed about the centre of mass [11,24].

**2.2 Bipedal Walking.**

In the context of biped robots, walking can be considered as a repetition of one-step motion [10,25]. The motion cycle can be divided into the single support phase and the double support phase just as shown in Figure 4. Single support refers to the fact that one foot holds the robot's weight while the other foot is moving forward on the air. The hip moves parallel to the ground keeping a constant height as shown in Figure 4.

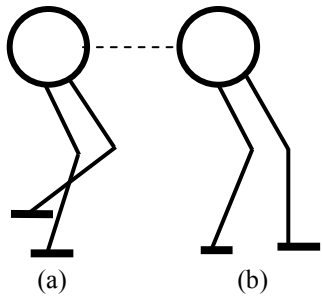

**Fig. 4.** Walking phases, a) single support, b) double support.

In the case of human walking, the orientation of the swing foot changes in order to reduce the impact between the foot and the ground. For robotic bipedal walking, this impact can be reduced by the z-directional motions of the swing foot and body and therefore the relative foot position and the ground must be known at all times. This fact indeed implies a full feedback and sensor data integration design considering that terrain conditions may change [10, 11, 12, 13].

## 3.  Impact model and velocity discontinuity.

This section describes the effects in terminal velocities in the foot-ground impact. This condition occurs when the swing leg touches the walking surface. Let $Q_s$ be the *N*-dimensional configuration space of the robot when the stance leg end is acting as a pivot and let $q_s = (q_1, \ldots, q_N) \in Q_s$ be a set of generalized coordinates. The swing phase model corresponds to an open kinematic chain. Applying the method of Lagrange (see [26]), the model is written in the form

$$D_s(q_s)\ddot{q}_s + C_s(q_s,\dot{q}_s)\dot{q}_s + G_s(q_s) = B_s(q_s)u \tag{4}$$

The matrix $D_s$ being the inertia matrix, $C_s$ being the Coriolis matrix, $G_s$ representing the gravity vector and $B_s$ mapping the joint torques to generalized forces. Expression $u = (u_1,\ldots,u_N) \in \mathbf{R}^N$ holds the torque being applied to each joint $i$ ($i \in (1,\ldots,N)$). It is clear so far that not all configurations of the model are physically compatible to the single support phase concept of walking as presented before. For example, all points of the robot should be above the walking surface excluding the end of the stance leg of course. In addition, there are some other kinematic constraints [27] that also must be considered.

The impact event is very short [28], so the ground reaction forces may be replaced by an impulse. The impact model therefore involves the reaction forces at the leg's end and requires the model of the biped robot [18,26]. Let $q_s$ be the generalized coordinates for the single support model and $\mathbf{c} = (c_x,c_y,c_z)$ the Cartesian coordinates of some fixed point mass on the robot. By using the generalized coordinates $q_e = (q_s,\mathbf{c})$, the Lagrange method yields

$$D_e(q_e)\ddot{q}_e + C_e(q_e,\dot{q}_s)\dot{q}_e + G_e(q_e) = B_e(q_e)u + \delta F_R \tag{5}$$

with $\delta F_R$ representing the reaction force acting on the robot due to the contact between the swing leg's end and the ground. According to [18], that results in a discontinuity for the velocity components of the biped robot. This will therefore be a new initial condition from which the single support model would evolve until the next impact appears as follows:

$$\dot{q}_e^+ = \Delta(\dot{q}_e^-), \tag{6}$$

with $\dot{q}_e^+$ and $\dot{q}_e^-$ being the velocities values just after and just before the impact respectively. Considering no rebounds, the impact can thus be modeled as in [18] as follows:

$$D_e(q_e^+)\dot{q}_e^+ - D_e(q_e^-)\dot{q}_e^- = F_R, \tag{7}$$

According to Equation 7, the value of the reaction force $F_R$ (impact) decreases when the difference among $\dot{q}_e^+$ and $\dot{q}_e^-$ is minimized. Such condition may be induced, according to Equation 6, if the trajectory speed $\dot{q}_e^-$ is minimized before impacting the ground [18].

## 4. Polynomial Trajectory Algorithm (PTA).

The position of the robot is controlled with respect to the frontal plane by motors M1, M6, M5 and M10 (See Figure 2(a)). The walking sequence of a biped robot can thus be determined by computing the hip and swing foot trajectories in the sagittal and frontal plane [29,30]. For the sagittal case, the servo control system drives motors M7, M8 and M9 for the left leg and M2, M3 and M4 for the right leg (Fig. 2(b)). In this work, the robot´s stability was achieved by applying the ZMP criteria, while Hermite spline interpolation is used to generate the walking trajectories as explained below.

**4.1 Hermite spline interpolation**

Cubic polynomials offer a reasonable balance between interpolation smoothness and computation speed. In comparison to polynomials of higher degree, the cubic splines require less calculations and smaller storage spaces, besides they are more stable.

A Hermite spline [31] is a cubic polynomial interpolation in segments with adjustable derivatives (velocities) at each control point. Opposite to natural cubic splines, Hermitian allows to define the segment locally because each part of the curve only depends on the conditions of its initial and final points. This is an important property to generate robot's trajectories, since it allows diminishing the speed while arriving to the final point, reducing the characteristic impact as the leg contacts the ground (see section 3).

If $P(u)$ represents a cubic parametric function that interpolates values between two control points (see Fig. 5), then the conditions to define the Hermite curve are:

$$\begin{aligned} P(0) &= P_s \\ P(1) &= P_e \\ P'(0) &= v_s \\ P'(1) &= v_e \end{aligned} \qquad (8)$$

with $v_s$ and $v_e$ expressing the initial and final velocity of the segment respectively.

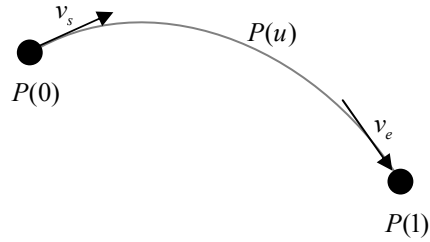

**Fig. 5.** Interpolation values between two control points.

The Hermite curve is thus defined as:

$$P(u) = au^3 + bu^2 + cu + d \qquad 0 \leq u \leq 1 \tag{9}$$

with *a, b, c* and *d* being the function parameters. This expression and its derivative can be expressed in matrix form yielding:

$$P(u) = \begin{bmatrix} u^3 & u^2 & u & 1 \end{bmatrix} \cdot \begin{bmatrix} a \\ b \\ c \\ d \end{bmatrix} \qquad P'(u) = \begin{bmatrix} 3u^2 & 2u & 1 & 0 \end{bmatrix} \cdot \begin{bmatrix} a \\ b \\ c \\ d \end{bmatrix} \tag{10}$$

By substituting the values of the initial and final points $P_s$ and $P_e$, as well as their velocities $v_s$ and $v_e$, the expression becomes

$$\begin{bmatrix} a \\ b \\ c \\ d \end{bmatrix} = \begin{bmatrix} 2 & -2 & 1 & 1 \\ -3 & 3 & -2 & -1 \\ 0 & 0 & 1 & 0 \\ 1 & 0 & 0 & 0 \end{bmatrix} \cdot \begin{bmatrix} P_s \\ P_e \\ v_s \\ v_e \end{bmatrix} \tag{11}$$

Now by replacing values *a, b, c* and *d* in Equation 9 and re-arranging the final polynomial emerges as follows:

$$P(u) = P_s(2u^3 - 3u^2 + 1) + P_e(-2u^3 + 3u^2) + v_s(u^3 - 2u^2 + u) - v_e(u^3 - u^2) \tag{12}$$

**4.2 Trajectory generation**

One walking motion can be considered as a repetition of one-step motion being repeated within a $T_s$ period. In order to achieve robust walking, the impact with the walking surface at the end of the single support phase should be executed smoothly by reaching zero velocity at the very contact spot. The walking sequence can thus be determined by only computing the trajectory of the hip and the swing foot and using the inverse kinematics to generate the trajectory for each independent joint in the biped structure.

**4.3 Hip trajectory in sagittal plane**

Hip trajectory can be generated using the Hermite spline polynomial algorithm considering that the initial and final states (position and velocities) are known from the single support phase. Figure 6 shows the initial position defined by $[x_{hs}, z_{hs}]$ and the final position represented by $[x_{he}, z_{he}]$. The initial velocity $[v_{xhs}, v_{zhs}]$ (produced when the robot leaves the initial position) is also specified in the trajectory model. The same applies for the final velocity $[v_{xhe}, v_{zhe}]$, exhibited when the robot contacts the walking surface.

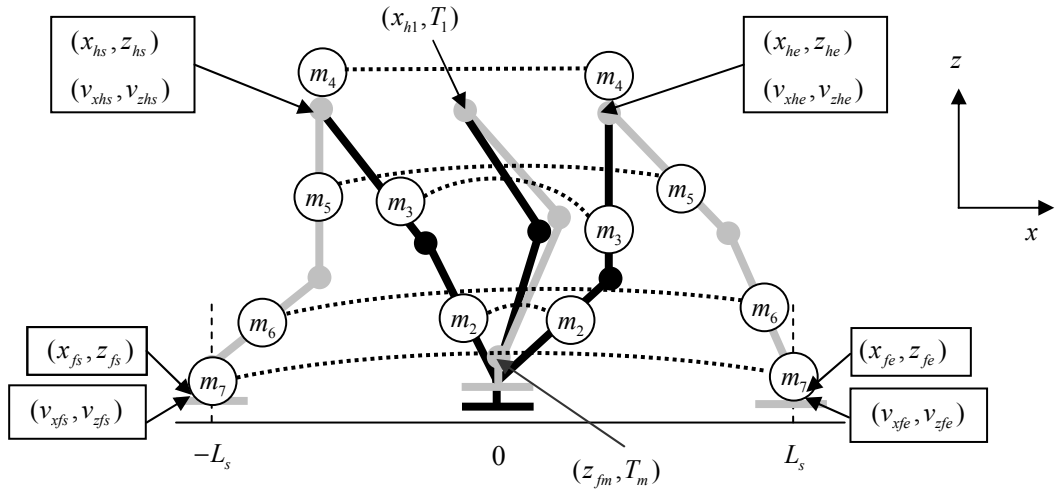

**Fig. 6.** Hip and swing foot trajectories in sagittal plane.

The initial and final positions for the cubic trajectory in *z* (the $z_h(t)$ direction), can be expressed as follows:

$$z_h(t) = \begin{cases} z_{hs} & \text{if} \quad t = kT \\ z_{he} & \text{if} \quad t = kT + T_s \end{cases} \tag{13}$$

with *T* being the period for the robot's step and $T_s$ the period in single support phase.

$$\dot{z}_h(t) = \begin{cases} v_{zhs} & \text{if} \quad t = kT \\ v_{zhe} & \text{if} \quad t = kT + T_s \end{cases} \tag{14}$$

Considering Eq. 12 the cubic polynomial can be defined as:

$$z_h(t) = z_{hs} + v_{zhs}(t - kT) + \frac{3(z_{he} - z_{hs}) - 2v_{zhs}T_s - v_{zhe}T_s}{T_s^2}(t - kT)^2 \\ + \frac{2(z_{hs} - z_{he}) + (v_{zhs} + v_{zhe})T_s}{T_s^3}(t - kT)^3 \qquad kT < t \leq kT + T_s \tag{15}$$

$x_h(t)$ is divided into two parts: from $x_h(KT)$ to $x_h(kT+T_1)$ and from $x_h(kT+T_1)$ to $x_h(kT+T_p)$. The definition for $x_h(t)$ can be summarized as follows:

$$\begin{cases} x_h(t) = x_{hs} & t = kT \\ x_h(t) = x_{h1} & t = kT + T_1 \\ x_h(t) = x_{he} & t = kT + T_s \\ \dot{x}_h(t) = v_{xhs} & t = kT \\ \dot{x}_h(t^-) = \dot{x}_h(t^+) & t = kT + T_1 \\ \dot{x}_h(t) = v_{xhe} & t = kT + T_s \\ \ddot{x}_h(t) = a_0 & t = kT \end{cases} \tag{16}$$

with $a_o$ being pre-specified to satisfy the initial condition of acceleration. The cubic polynomial trajectory can thus be calculated using (15), yielding:

$$x_h(t) = \begin{cases} x_{hs} + v_{xhs}(t-kT) + \frac{1}{2}a_0(t-kT)^2 \\ + \dfrac{(x_{h1} - x_{hs} - v_{xhs}T_1 - \frac{1}{2}a_0T_1^2)(t-kT)^3}{T_1^3} & kT < t \leq kT + T_1 \\ x_{h1} + v_{xh1}(t-kT-T_1) \\ + \dfrac{(3(x_{he} - x_{h1}) - 2v_{xh1}(T_p - T_1))(t-kT-T_1)^2}{(T_p - T_1)^2} \\ + \dfrac{(2(x_{h1} - x_{he}) + (v_{xh1} + v_{x2})(T_p - T_1))(t-kT-T_1)^3}{(T_p - T_1)^3} & kT + T_1 < t \leq kT + T_s \end{cases} \quad (17)$$

### 4.4 Swing foot trajectory in sagittal plane.

Hermite spline polynomial interpolation is used to generate the foot trajectory in the single support phase. In order to assure a smooth transition, velocities ($(v_{xhs}, v_{zhs})$ and $(v_{xfe}, v_{zfe})$) should be defined near zero. The final foot position (Figure 6) representing target positions and velocities can thus be obtained following:

$$x_f(t) = \begin{cases} x_f(t) = x_{fs} & t = kT \\ x_f(t) = x_{fe} & t = kT + T_s \\ \dot{x}_f(t) = 0 & t = kT \\ \dot{x}_f(t) = 0 & t = kT + T_s \end{cases} \quad (18)$$

$$z_f(t) = \begin{cases} z_f(t) = z_{fs} & t = kT \\ z_f(t) = z_{fe} & t = kT + T_s \\ \dot{z}_f(t) = 0 & t = kT \\ \dot{z}_f(t) = 0 & t = kT + T_s \end{cases} \quad (19)$$

From the initial and the final positions in *x* and *z* axis, a smooth trajectory can be generated by the Hermite spline interpolation yielding:
for *x$_f$(t)*:

$$x_f(t) = x_{fs} + 3(x_{fe} - x_{fs})\frac{(t-kT)^2}{T_s^2} - 2(x_{fe} - x_{fs})\frac{(t-kT)^3}{T_s^3} \quad kT < t \leq kT + T_s \quad (20)$$

and for $z_f(t)$:

$$\begin{cases} z_{fs} + 3(z_{fm} - z_{fs})\dfrac{(t-kT)^2}{T_m^2} - 2(z_{fm} - z_{fs})\dfrac{(t-kT)^3}{T_m^3} & kT < t \leq kT + T_m \\ z_{fm} + 3(z_{fe} - z_{fm})\dfrac{(t-kT-T_m)^2}{(T_s - T_m)^2} - 2(z_{fe} - z_{fm})\dfrac{(t-kT-T_m)^3}{(T_s - T_m)^3} & kT + T_m < t \leq kT + T_s \end{cases} \quad (21)$$

The hip and knee positions required to produce appropriate movements for each leg can thus be calculated using the inverse kinematics of the robot's structure.

**4.5 Hip trajectory in frontal plane**

The hip trajectory can also be generated by the Hermite spline polynomial algorithm. However, all movements in the frontal plane should be performed within a cycle (see Figure 7). The hip moves from the initial position ($y_{hs}$) to the maximum allowed displacement ($y_{he}$), with an initial velocity ($v_{yhs}$). Such movement must be completed within the half of the walking period ($T_2 = T_s/2$). From that position ($y_{he}$), the hip moves again to its initial position ($y_{hs}$), by executing a very smooth trajectory. It allows diminishing the speed while arriving to the final point by reducing the characteristic impact on the leg at ground contact.

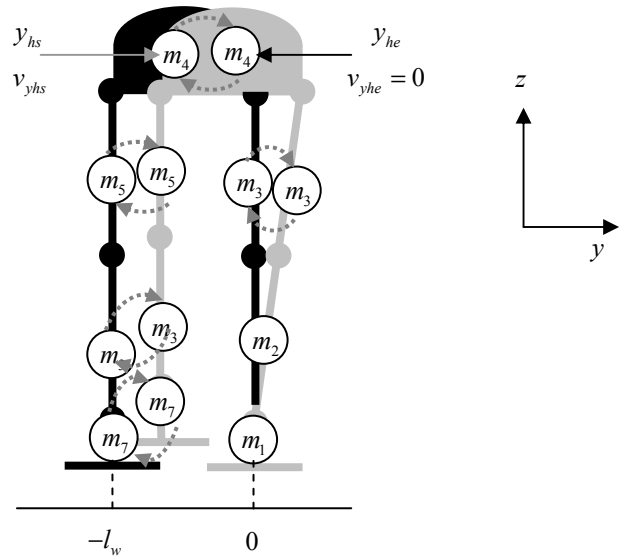

**Fig. 7.** Hip and swing foot trajectories in the frontal plane.

From initial and final positions in *y*, a smooth trajectory can be generated by the Hermite spline interpolation yielding:

for $kT < t \leq T_2$

$$y_h(t) = y_{hs} + v_{yhs}(t-kT) + \frac{3(y_{he}-y_{hs})-2v_{yhs}T_2}{T_2^2}(t-kT)^2 + \frac{2((y_{hs}-y_{he})+v_{yhs}T_2}{T_2^3}(t-kT)^3 \tag{22}$$

for $T_2 < t \leq T_s$

$$y_h(t) = y_{he} + \frac{3(y_{hs}-y_{he})}{(T_s-T_2)^2}(t-kT)^2 - \frac{2((y_{hs}-y_{he}))}{(T_s-T_2)^3}(t-kT)^3 \tag{23}$$

### 4.6 Determination of the algorithm parameters

Tuning $v_{xhs}$, $T_l$, $T_m$, $z_{fm}$ and $v_{yhs}$ has immediate impact on the smoothness of the trajectory with respect to velocities and accelerations. Recalling that the link's speed at the end of the single support phase should be approaching zero in order to assure a smooth contact with the floor surface, a proper selection is relevant for the overall performance. If $T_l = T_m = T_v$ then only $T_v$ needs to be fixed. If $T_v$ is close to $T_s/2$ then a robust walking trajectory with smooth impact transition may be obtained. In turn, this feature allows robustness against disturbances either from non-modeled dynamics of the biped robot or from the walking surface.

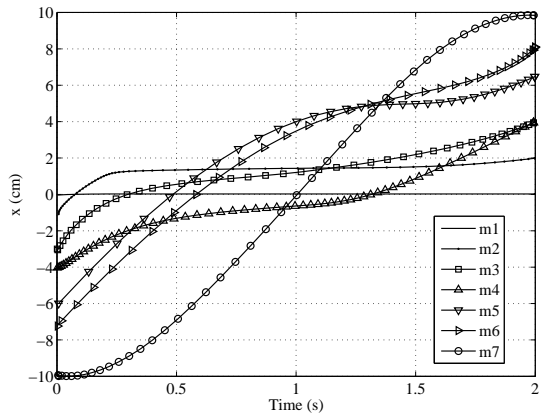
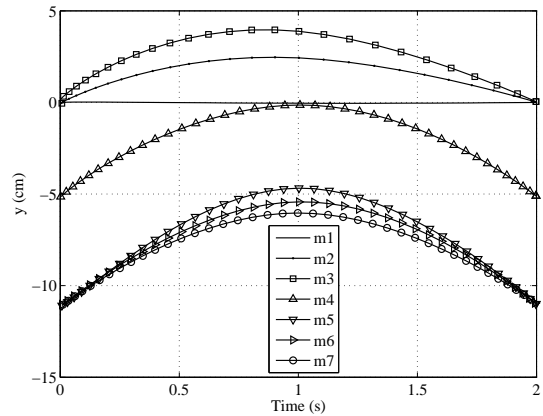

(a) (b)

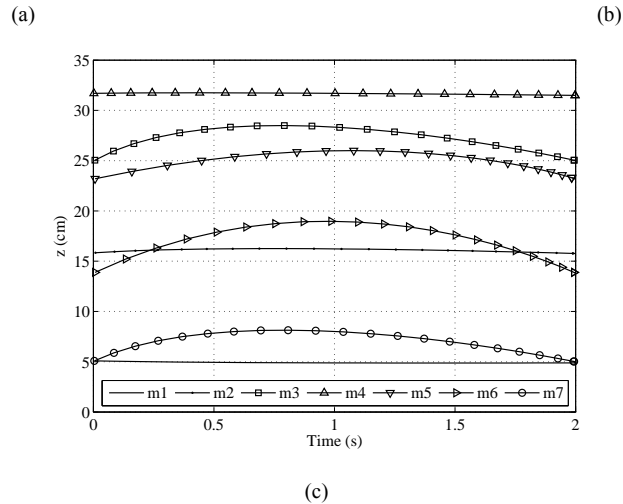

(c)

**Fig. 8.** Walking motion, (a) *x* trajectory, (b) *y* trajectory and (c) *z* trajectory.

## 5. Walking motions

Motion can be thus generated according the PTA algorithm following Equations 13-23. The trajectory for each point mass ($m_i$) is shown in Figure 8. A step period of period $T_s = 2$ s and step length $L_s = 10$ cm are considered. Figure 9 shows the motion on the saggital plane for each link, considering the trajectory of each mass point. The evolution of the trajectories may be simulated considering the "Dany Walker" model according to Table 1. It can be seen that the trajectories shown in Figure 9 match those calculated for PTA as it is shown by Figure 8. All tests consider a flat walking surface.

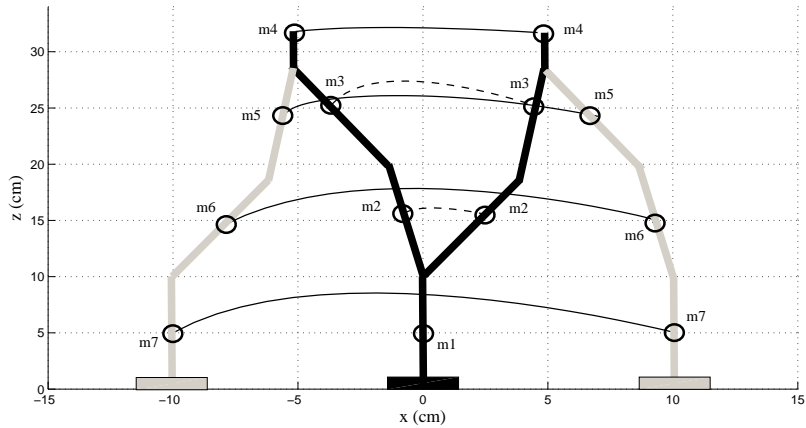

**Fig. 9.** Walking motion trajectory in the saggital plane.

Figure 10 shows the trajectories for the ZMP as they are generated by PTA for real-time experiments in the robot. The values of the ZMP were calculated from a triangular arrangement of force sensors located at each foot of the "Dany Walker" robot [19,21].

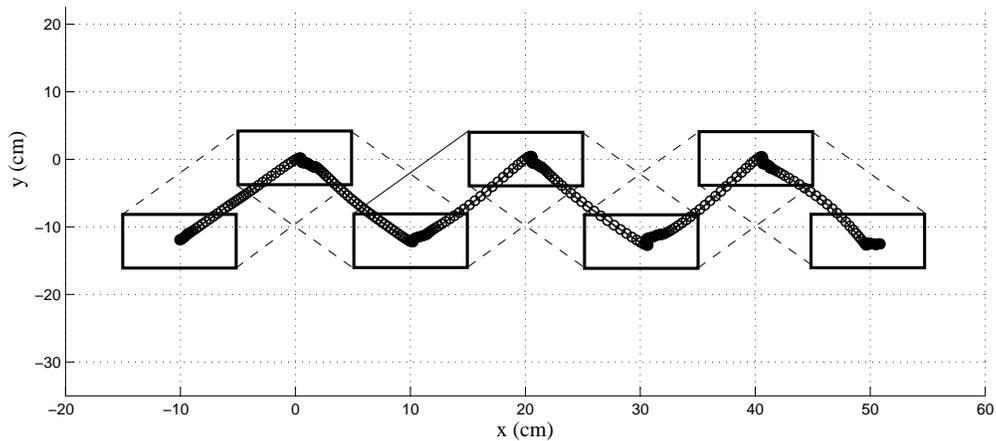

**Fig. 10.** ZMP trajectory using PTA.

## 5.1 Experiments

In order to demonstrate the effectiveness of the PTA algorithm, the step motion is tested under the assumption that the contact between the swing foot and the ground happens on uneven terrain. Two experiments are prepared. First, the impact is simulated over a variable-height surface in order to test the PTA performance. The experiment includes a comparison between PTA and other trajectory generation methods. The second experiment is a real-time test of the resulting PTA's trajectories. Small wooden pieces are arranged over the robot's path. They are lightly bigger that the size of the robot's foot, aiming to create a changing relationship between the swinging foot and the floor. In order to compare the results, the same experiment was proved with the broadly well-known approach (according to references [33-37] found in the literature) presented in [10].

*5.1.1 Simulation of the impact on the floor.*

The first experiment employs the "Dany Walker" biped robot. The kinematic and dynamic models of the robot are presented in [20] and [19]. The impact model proposed in [18] is used to test the actual ability of PTA to achieve a stable walking within a variable-height terrain whose attitude varies from 0 to 1cm at about 6cms away from the starting point (see Figure 11). The simulation considers $T_s = 5$ s and $L_s = 10$ cm. Figure 11 shows the trajectories produced by PTA and the zoom of the transitory produced over the mass 6, at the impact instant. After the simulation, it can be assumed that the speed of $m_7$ ($\dot{m}_7^-$) was 3.17cm/s producing an equivalent force of $F_R = 0.11N$ at the impact.

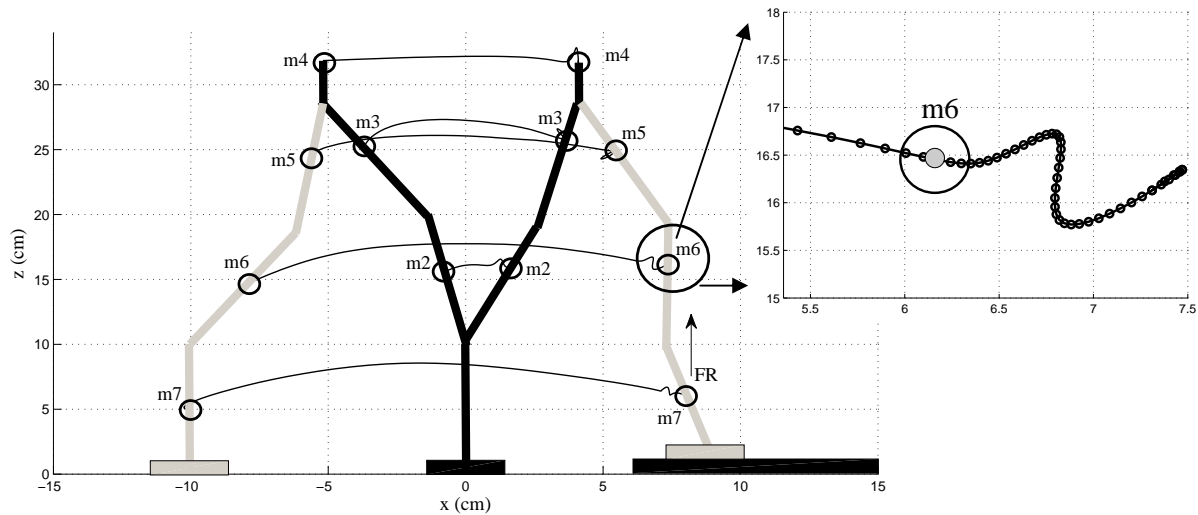

**Fig. 11.** Walking PTA trajectory in the saggital plane under variable terrain conditions. (Upper right) A zoom view of the transient response of the mass m6.

Table 2 shows the performance of the PTA as it is compared to other IPM trajectory generation algorithms such as the Ohishi's method [11], the Kajita's algorithm [17], Park's contribution in [10] and the Van der Pol oscillator as it is applied in [15]. The experiment employed the same parameters reported by the authors in order to ease the comparison to the "Dany Walker" robot.

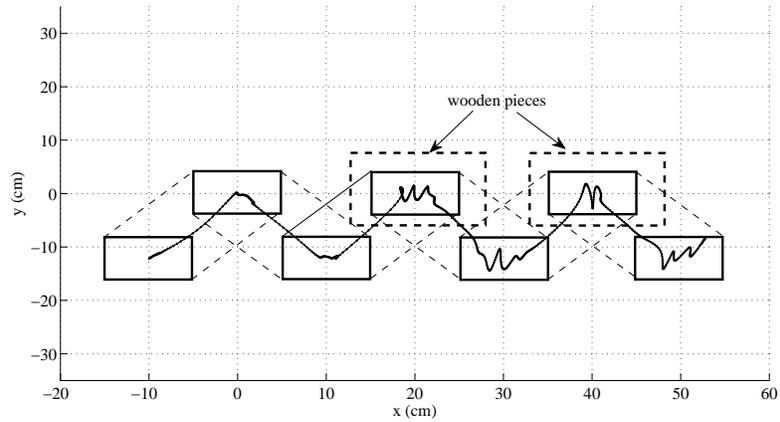

(a)

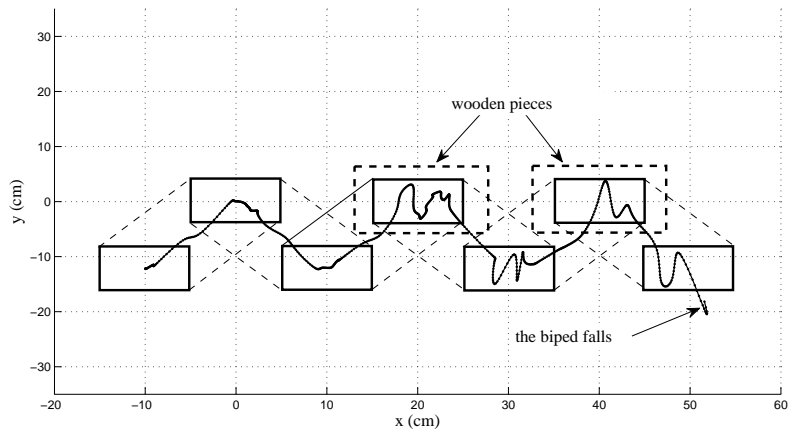

(b)

**Fig. 12.** Second experiment: real-time impacts on the ground. The ZMP trajectories are shown while the robot walks over an uneven terrain.

(a) Shows the PTA operation while (b) shows the IPM Park's algorithm.

| Algorithms | Velocity of $m_7$ ($\dot{m}_7^-$) before the impact | $F_R$ |
|---|---|---|
| PTA | 3.17cm/s | 0.11$N$ |
| IPM Ohishi version | 9.51cm/s | 0.821$N$ |
| IPM Kajita version | 10.44cm/s | 0.902$N$ |
| IPM Park version | 8.82cm/s | 0.711$N$ |
| Van der Pol oscillator | 13.71cm/s | 1.05$N$ |

**Table 2.** Results of the first experiment: PTA vs. other trajectory generation algorithms under changing terrain conditions.

The PTA reduces the speed and therefore produces a small reaction force $F_R$ at the impact (near to zero), compared to other algorithms based on IPM or on Van der Pol oscillators which hold the highest velocity at the impact with 13.71cm/s. The IPM algorithms adjust the trajectories according to accelerations yielding higher velocities before the impact while the Van de Pol-based algorithm keeps constant speeds even when the surface's height is increased. The value of the reaction force $F_R$ (impact) decreases when the velocity $\dot{m}_7^-$ is minimized.

*5.1.2 Impacts on the real-time walking*
This experiment aims to set a new walking surface by scattering wooden pieces of same dimensions of the biped's foot but being 2 cm taller than them at each side. Obstacles begin at about 15 cm away from the initial position and they end at about 50 cm at the robot's front flank. Figure 12(a) shows the results of PTA, while Figure 12(b) presents the performance of the IPM Park algorithm [10] over the same experimental setting. The ZMP trajectory presented in Figure 12(a) demonstrates that the ZMP is always kept inside the foot support area using PTA, despite disturbances related to the terrain type. Figure 12(b) shows the ZMP trajectory is not always kept within the foot support area. Therefore, the robot is only able to complete two steps before falling down at an uneven terrain.

The ZMP readings were estimated from the triangular arrangement of force sensors at each foot [19, 21]. The generation of both trajectories considers $T_s$ = 5 sec and $L_s$ = 10 cm. The PTA algorithm was programmed using Visual C++ and was tested on the "Dany Walker" biped robot [19]. Figure 13 shows the real-time walking sequence.

Table 3 shows the average values of the reaction force for 20 cycles. In order to measure the reaction force, a force-voltage relationship [38] is used. Only left foot impacts have been considered.

| Algorithms | $\bar{F}_R$ |
|---|---|
| IPM Park version | 0.717N |
| PTA | 0.201$N$ |

**Table 3.** Results of the second experiment: PTA vs. IPM Park algorithms under changing terrain conditions.

**5.2 Computational Cost**

This section compares the computational cost of PTA to other generation algorithms such as Hermitian polynomials and B-Splines. The time required to calculate the solution and the required storage space are considered as comparison indexes. All systems are compiled using the CCS PIC-C® 4.068 compiler aiming for the PIC18F4550 microcontroller platform.

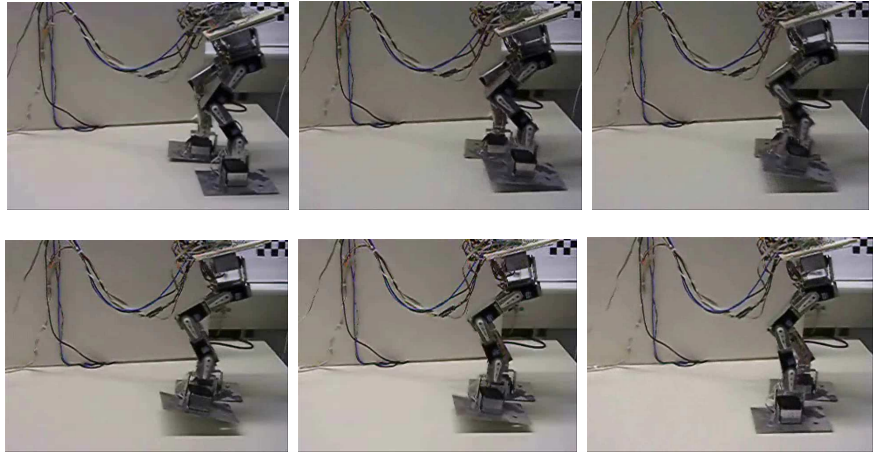

**Fig. 13.** Walk sequence using the TPA algorithm.

The experiment considers a complete step trajectory and its computation. Table 4 shows results considering several methods such as PTA, IPM Ohishi method [11], IPM Kajita algorithm [17], IPM Park procedure [10] and the Van der Pol oscillator algorithm [15].

Table 3 shows that only the IPM Park version can actually generate acceptable gaits, considering the modest computer resources of the robot. Other methods cannot generate speeds smaller than 4 seconds which poses a serious difficulty to perform control on the system by considering the signal sensors feedback and the calculus of the ZMP. Besides, to implement the IPM-based algorithms (Ohishi, and Kajita) and the Van del pol oscillator, it was necessary to expand the memory of the host computer system because one numerical method is required to solve the pendulum-based system for IPM methods or to generate the oscillation trajectories for the Van der Pol method.

| Algorithm | Calculus time | Storage space |
|---|---|---|
| PTA | 0.4s | 86 Bytes |
| IPM Ohishi version | 4s | 415 Bytes |
| IPM Kajita version | 4s | 415 Bytes |
| IPM Park version | 2s | 211 Bytes |
| Van der Pol oscillator | 12s | 518 Bytes |

**Table 4.** PTA computational cost vs other trajectory generation algorithms.

The B-splines class is a very well-known tool which following the Hermitian polynomials method, may be generated by simply approaching their control points (allowing to configure positions and speeds). However the B-spline allows a better control of the intermediate points of the trajectory, in such a way that they can be modified locally without a substantial change in the complete trajectory. In the case of the generation of trajectories for biped robots, the intermediate points are not modified because the final point of the trajectory is what really matters. Under this fact, the use of B-splines would not represent any advantage in comparison to the Hermitian polynomials. However its use would imply a bigger computational cost, since its calculation involves the use of more complex algorithms [32].

## 6. Conclusions

In this paper, the Polynomial Trajectory Algorithm (PTA) was proposed as a simple algorithm to generate successful walking trajectories for biped robots. The equations used by PTA are obtained from the Hermite spline interpolation of initial and final conditions. The algorithm is able to generate smooth impacts against the ground when the robot changes between different walking phases. It was demonstrated that these trajectories reduce the trend for falling down when the terrain condition changes. Several joint trajectories for walking motion on the 3D robot "Dany walker" were tested and compared to the results offered by other IPM methods. The results showed that by using PTA the impact effect is minimized. The paper also studies the ZMP trajectory produced by PTA and the IPM Park's method. The results showed that applying the IPM algorithm, the robot is able to complete only two steps within an uneven surface before falling down. PTA is shown to be simple and suitable to generate trajectories for bipedal walking in real time despite the use of modest computing resources such as a commercial embedded microcontroller. Since PTA only implements Equations 13- 23, it does demand modest computing resources allowing its application to other humanoid robot platforms.

## References

**[1]** R. Goddard, Y. Zheng and H. Hemami, Control of the heel-off to toe-off motion of a dynamic biped gait, *IEEE Trans. Syst. Man. Cayb., vol.22*, no. 1, 1992.

**[2]** N. Kanehira, T. Kawasaki, S. Ohta, T. Isozumi, T. Kawada, F. Kanehiro, S. Kajita & K. Kaneko, Design and experiments of advanced module (HRPL-2L) for humanoid robot (HRP-2) development, Proc. 2002, IEEE-RSJ *Int. Conf. Intell. Rob. Sys. EPFL*, Lausanne, Switzerland, 2002, 2455-2460.

**[3]** A. Konno, N. Kato, S. Shirata, T. Furuta & M. Uchiyama, Development of a light-weight biped humanoid robot in Proc. 2000 *IEEE-RSJ Int. Con. Intell. Rob. Sys.*, 2000, 1565-1570.

**[4]** M. Vukobratović, B. Borovac, D. Surla, D. Stokic, Biped Locomotion. *Dynamics, Stability, Control and Application*, Springer-Verlag, London, UK, 1990.

**[5]** Amos Albert, Wilfried Gerth, Analytic path planning algorithms for bipedal robots without a trunk, *Journal of Intelligent and Robotic Systems 36* (2003) 109–127.

**[6]** Atsuo Takanishi, Mamoru Tochizawa, Hideyuki Karaki, Ichiro Kato, Dynamic Biped Walking Stabilized with Optimal Trunk and Waist Motion, in: *Proceedings of the IEEE/RSJ International Conference on Intelligent Robots and Systems*, Sept. 1989, pp. 187–192.

**[7]** Jong Hyeon Park, Fuzzy-logic zero-moment-point trajectory generation for reduced trunk motions of biped robots, *Fuzzy Sets and Systems 134* (2003) 189–203.

**[8]** Taesi Ha, Chong-Ho Choi, An effective trajectory generation method for bipedal walking. *Robotic and Autonomus Systems 55* (2007) 795-810.

**[9]** R. Héliot, B. Espiau, Online generation of cyclic leg trajectories synchonized with sensor measurement. *Robotics and Autonomus Systems*. In press. (2007)

**[10]** Jong H. Park, Kyoung D. Kim, Biped Robot Walking Using Gravity Compensated Inverted Pendulum Mode and Computed Torque Control, in: *Proceedings of the IEEE International Conference on Robotics and*

*Automation*, vol. 4, May 1998, pp. 3528–3533.

**[11]** K. Ohishi, K. Majima, T. Fukunaga and T. Miyazaki. Gait Control of a Biped Robot Based on Kinematics and Motion Description in Cartesian Space. *23 International Conference on Industrial Electronics, Control and Instrumentation (IECON 97)*, pp 1317-1322.

**[12]** S. Kajita and K. Tani. Experimental study of biped dynamic walking. *IEEE Control Systems Magazine*, 16(1):13–9, February 1996.

**[13]** S. Kajita, T. Yamaura, and A. Kobayashi. Dynamic walking control of biped robot along a potential energy conserving orbit. *IEEE Transactions on Robotics and Automation*, 8(4):431–37, August 1992.

**[14]** A. A. Grishin, A. M. Formal'sky, A. V. Lensky, and S. V. Zhitomirsky. Dynamical walking of a vehicle with two legs controlled by two drives. *International Journal of Robotics Research*, 13(2):137–47, 1994.

**[15]** R. Katoh and M. Mori. Control method of biped locomotion giving asymptotic stability of trajectory. *Automatica*, 20(4):405–14, 1984.

**[16]** J. Furusho and M. Masubuchi. Control of a dynamical biped locomotion system for steady walking. *Journal of Dynamic Systems, Measurement and Control*, 108:111–8, 1986.

**[17]** S. Kajita, F. Kanehiro, K. Kaneko, K. Fijiware, K. Harada, K. Yokoi and H. Hirukawa. Biped Walking Pattern generation by using Preview Control of Zero-Moment Point. *IEEE International Conference on Robotics & Automation, ,* pp 1620-1626. 2003.

**[18]** Y. Hürmüzlü and D. B. Marghitu. Rigid body collisions of planar kinematic chains with multiple contact points. *International Journal of Robotics Research*, 13(1):82–92, 1994.

**[19]** Zaldivar-Navarro, Daniel, 2006, "A Biped Robot Design", PhD Thesis, FU Berlin, Fachbereich Mathematik u. Informatik, Freie Universität Berlin, available on-line http://www.diss.fu-berlin.de/2006/664/indexe.html.

**[20]** Erik V. Cuevas, Daniel Zaldívar, and Raúl Rojas, Bipedal robot description, *Technical Report B-03-19, Freie Universität Berlin, Fachbereich Mathematik und Informatik*, Berlin, Germany,   2004.

**[21]** Erbatur, K., Okazaki, A., Obiya, K., Takahashi, T., and Kawamura, A. A study on the zero moment point measurement for biped walking robots. *7th International Workshop on Advanced Motion Control, IEEE*. (2002).

**[22]** Qiang Huang, K. Yokoi, S. Kajita, K. Kaneko, H. Arai, N. Koyachi, K. Tanie, Planning walking patterns for a biped robot, *IEEE Transactions on Robotics and Automation,* 17 (2001) 280–289.

**[23]** Duško Katić, Miomir Vukobratović, Survey of intelligent control techniques for humanoid robots, *Journal of Intelligent and Robotic Systems 37* (2003) 117–141.

**[24]** Erbatur, K., Okazaki, A., Obiya, K., Takahashi, T., and Kawamura, A.. A study on the zero moment point measurement for biped walking robots. *7th International Workshop on Advanced Motion Control,* 2002.

**[25]** A. Yonemura, Y. Nakajima, A.R. Hirakawa, A. Kawamura, Experimental approach for the biped walking

robot MARI-1, in: 6th *InternationalWorkshop on Advanced Motion Control*, March–April 2000, pp. 548–553.

**[26]** E. Westervelt, J. Grizzle, C. Chevallerau, J Choi and B. Morris. *Feedback Control of Dynamic Bipedal Robot Locomotion*. CRC Press, NW, US, 2007.

**[27]** B. Brogliato. Nonsmooth Impact Dynamics: Models, Dynamics and Control, volume 220 of *Lecture Notes in Control and Information Sciences*. Springer, London, 1996.

**[28]** Y. Fujimoto and A. Kawamura. Simulation of an autonomous biped walking robot including environmental force interaction. *IEEE Robotics and Automation Magazine*, pages 33–42, June 1998.

**[29]** A. Takanishi, M. Ishida, Y. Yamazaki & I. Kato, The realization of dynamic walking robot WL-10RD*, in Proc. 1985*, *Int. Conf. Advanced Robotics* , 1985, 459-466.

**[30]** K. Hiria, M Hirose, Y. Haikawa, & T.Takenaka, The development of Honda Humanoid robot, *in   IEEE Int. Conf. Rob. and Aut.* , 1998, 1321-132.

**[31]** M. Fedoryuk, Hermite polynomials, *Encyclopaedia of Mathematics*, Kluwer Academic Publishers, ISBN 978-155608010, 2001.

**[32]** T. Popiel, On parametric smoothness of generalised B-spline curves, *Computer Aided Geometric Design*, Volume 23, Issue 8, November 2006, Pages 655-668.

**[33]** K. Nishiwaki and S. Kagami, Online Walking Control for Humanoids with Short Cycle Pattern Generation, *The international Journal of Robotics Research*, 28 (2009) 729–742.

**[34]** S. Bououden, F. Abdessemed and B. Abderraouf, Control of a Bipedal Walking Robot Using a Fuzzy Precompensator, *Agent and Multi-Agent Systems: Technologies and Applications*, A. Håkansson et. Al (Eds), Springer-Verlag, Berlin 2009.

**[35]** C. Zeng, Y. Cheng, H. Liang, L. Dai and H. Liu, Research on the model of the inverted pendulum and its control based on biped robot, *IEEE Pacific-Asia Workshop on Computational and Industrial Application*, 2008.

**[36]** J. Park, Fuzzy-logic zero moment point trajectory generation for reduced trunk motions of biped robots, *Fuzzy Sets and Systems* 134, (2003), 189-203.

**[37]** S. Kajita, F. Kanehiro, K. Kaneko and K. Fujiwara, A real-time Pattern generation for a Biped Walking, *International Conference on Robotics & Automation*, 2002, Washington, DC.

**[38]** http://www.tekscan.com/flexiforce/flexiforce.html